\documentclass[conference]{IEEEtran}

\usepackage[T1]{fontenc}
\usepackage{times}
\usepackage[utf8]{inputenc}
\usepackage{amsmath,amssymb,bm}
\usepackage{graphicx}
\usepackage{booktabs}
\usepackage{xcolor}
\usepackage{multirow}
\usepackage{url}
\usepackage[numbers,sort&compress]{natbib}
\usepackage[bookmarks=true,colorlinks=true,allcolors=blue]{hyperref}

\newcommand{\q}{\bm{q}}
\newcommand{\Q}{\bm{Q}}

\begin{document}

\title{DemoBridge: A Simulation-in-the-Loop Toolkit for\\
Single-View Human Demonstration Retargeting}

\author{
\IEEEauthorblockN{
Zehao Wang\IEEEauthorrefmark{4},
Fabien Despinoy\IEEEauthorrefmark{2},
Sergey Zakharov\IEEEauthorrefmark{3},
Tinne Tuytelaars\IEEEauthorrefmark{4},
and Rahaf Aljundi\IEEEauthorrefmark{2}
}
\IEEEauthorblockA{
\IEEEauthorrefmark{4}KU Leuven
\IEEEauthorrefmark{2}Toyota Motor Europe
\IEEEauthorrefmark{3}Toyota Research Institute
}
}
\maketitle

\begin{abstract}
We present \textbf{DemoBridge}, an toolkit that turns a single-view RGB stereo recording of a human hand demonstration into an executable, physics-validated robot-arm trajectory. Retargeting across the embodiment gap is hard. A robot arm reaches a target with a long, articulated body whose links carry far more collision volume than a hand. Solving inverse kinematics (IK) for the mapped end-effector pose often yields no collision-free solution, and a trajectory imposes this at every waypoint. A single view adds noise, leaving the demonstrated reference inaccurate.
At the core of DemoBridge is a single collision-aware planner. It optimizes the whole joint trajectory at once, reasoning jointly over alternative grasp poses, whole-arm and grasped-object collision, and fidelity to the demonstrated path. A physics simulator runs in the loop. It validates each phase as it is produced and backtracks on failure, so a demonstration that cannot be reproduced as given is re-planned rather than discarded.
The resulting action sequence is dynamically stable and faithful to the demonstrated manipulation. It also doubles as a ready-to-use simulation rollout for policy learning. Grasp timing is inferred automatically, and the perception backends, robot, and pipeline stages are swappable from configuration.
We evaluate whole-pipeline retargeting on three real-demonstration tasks and the planner on a controlled synthetic benchmark. Our code is available at
\href{https://gitlab.kuleuven.be/u0123974/demo-bridge/}
{\texttt{gitlab.kuleuven.be/u0123974/demo-bridge}}.

\end{abstract}

\IEEEpeerreviewmaketitle

\section{Introduction}
\label{sec:intro}

Learning robot manipulation policies from demonstrations is effective, but collecting such demonstrations at scale remains difficult. Teleoperation~\citep{qin2022dexpilot} collects demonstrations by directly controlling the robot, which requires exclusive access to the platform and can interrupt ongoing deployment, such as on an assembly line. Motion capture, whether multi-camera rigs~\citep{wang2024hocap,wang2024dexcap} or instrumented gloves~\citep{liu2017glove}, needs a calibrated setup or worn sensors. Both confine demonstrations to a controlled environment. Ordinary single-view human video needs only one camera. A tool that turns such a recording into robot motion would let manipulation data be gathered passively and in place. It would also feed the real-to-sim-to-real and imitation methods that learn from human video~\citep{dan2025xsim,barcellona2024drema,lum2025crossinghuman,zhai2026imitatingworks}. The recorded hand motion must be transcribed into a trajectory a specific robot can physically execute, across the embodiment gap between a human hand and a robot arm. We provide the toolkit that performs this transcription, turning a single-view human recording into a robot-executable, physics-validated trajectory.

Two difficulties stand in the way, and they are coupled. The first is kinematic. A robot arm spans the workspace with an articulated chain of large collision volume. Even a single mapped end-effector target may have no collision-free inverse-kinematics (IK) solution. Retargeting a trajectory further compounds the problem, since such feasibility must be maintained at every waypoint along the demonstrated path. The second is perceptual. From a single view, hand keypoints and object poses jitter, drift, and are occluded at the moment of contact. The demonstrated reference is therefore inaccurate exactly where it matters most. Feasibility, contact, and reference fidelity jointly determine an executable motion, so they must be addressed together.

The systems closest to ours reach the demonstrated poses one at a time, by per-frame IK or by interpolating sparse keyframes~\citep{zhang2025robowheel,zhou2025yoto}. This is local: it reproduces each pose but reasons neither about whole-arm collision along the path nor about the noise in the reference. We take the opposite view. Instead of reaching poses one at a time, our planner optimizes a whole-arm trajectory that tracks the demonstrated path while keeping both the arm and the grasped object collision-free. A physics simulator validates every phase and triggers a re-plan when one fails, so a demonstration that cannot be reproduced as given is re-planned rather than discarded.

We instantiate this approach in \textbf{DemoBridge}, an toolkit for converting single-view human demonstrations into robot-executable trajectories. At its core, a retargeting solver jointly handles noisy visual references, robot feasibility, and collision avoidance, while a modular service layer connects perception, planning, and physics validation. Our contributions are:
\begin{itemize}
\item \textbf{A unified retargeting planner.} A single collision-aware trajectory optimizer couples grasp selection, whole-arm and grasped-object collision, and tracking of the demonstrated end-effector path. One objective covers a goal-only reach, a reference-following motion, and the transport of a grasped object.
\item \textbf{Simulation-in-the-loop, annotation-free retargeting.} Every phase is executed and validated in a physics simulator (Isaac Sim~\citep{isaaclab2023}) and re-planned on failure. Grasp and release timing are inferred from the hand--object relation rather than annotated. The emitted trajectory is dynamically consistent.
\item \textbf{A modular, configuration-driven toolkit.} Scene reconstruction, hand tracking, and object 6D pose tracking run as reusable services whose backends, robot, and pipeline stages swap from config without code changes.
\end{itemize}

The rest of the paper expands on this design: the modular architecture and backends (Sec.~\ref{sec:system},~\ref{sec:design}) and the retargeting solver (Sec.~\ref{sec:solver}): its event extraction, coordinator, and motion planner. We close with the evaluation (Sec.~\ref{sec:eval}): whole-pipeline success on three real demonstrations, and the planner on controlled synthetic scenes.

\begin{figure*}[t]
\centering
\includegraphics[width=\textwidth]{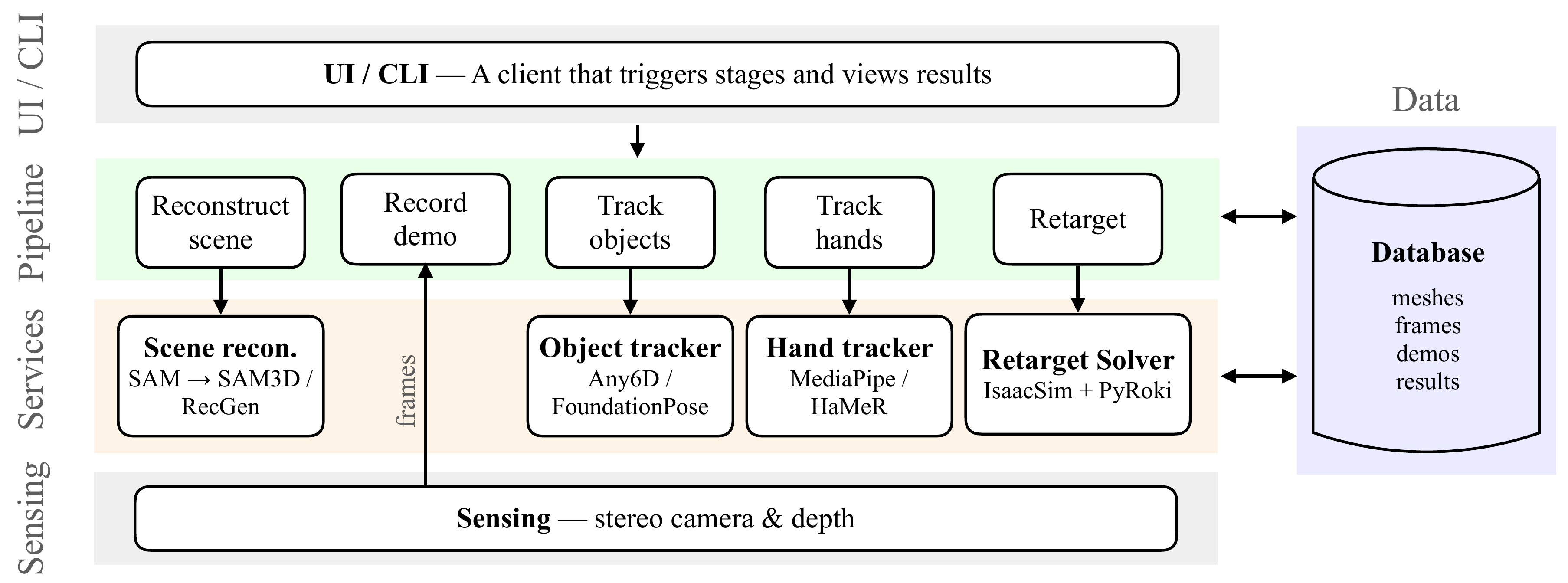}
\caption{DemoBridge's four-layer architecture. A thin UI/CLI triggers \emph{pipeline} stages; each stage calls a load-once \emph{service} and reads or writes a shared \emph{database}. The perception backends are swappable; the \textbf{Retarget} stage runs our simulation-in-the-loop solver, whose motion planner is a persistent service reused across a scene. The \emph{hardware} layer (stereo camera and depth) is reused from the host stack.}
\label{fig:arch}
\end{figure*}

\section{Related Work}
\label{sec:related}

\paragraph{Collecting manipulation demonstrations}
Scaling manipulation data has largely meant instrumenting the demonstrator, with multi-camera rigs that fit and retarget a parametric hand~\citep{wang2024hocap}, mocap gloves~\citep{liu2017glove,wang2024dexcap}, or teleoperation~\citep{qin2022dexpilot}. This raises cost and constrains where data can be collected. Retargeting from ordinary single-view human video removes this instrumentation, at the price of noisier, partially observed input.

\paragraph{Hand-to-robot retargeting}
Retargeting maps a demonstrated wrist or fingertip trajectory onto an end-effector target and realizes it on the robot~\citep{kim2025pyroki,curobo_v2}. Existing methods differ in how the target poses are produced and reached. The most direct methods reach the demonstrated poses individually by IK: RoboWheel~\citep{zhang2025robowheel} tracks the dense end-effector path with per-frame IK, and YOTO~\citep{zhou2025yoto} compresses the demonstration into sparse keyframes reached by IK interpolation. SPIDER~\citep{pan2025spider} retargets human hand motion to dexterous robot hands, refining in-hand contact dynamics by physics-based sampling. A generative line of work replaces the recorded demonstration with a synthesized one: FlowHOI~\citep{zeng2026flowhoi} generates hand--object interactions from a language and scene description and retargets them to a dexterous hand. Across these methods, the arm trajectory is never planned against the surrounding scene. They reach poses or contacts without reasoning about whole-arm collision along the path. Under a single view these concerns become coupled: the mapped pose may be infeasible and the reference is noisy, so they cannot be resolved in isolation. We address them jointly with an integrated, simulation-validated planner (Sec.~\ref{sec:solver}).

\section{System Overview}
\label{sec:system}

DemoBridge retargets a human hand demonstration to a robot arm in Isaac Sim~\citep{isaaclab2023}. A scene is reconstructed once and shared by many demonstrations; each demonstration is tracked and retargeted independently. The system has four replaceable, independently testable layers and a shared data store (Fig.~\ref{fig:arch}).

\noindent\textbf{UI / CLI.} A thin client that owns no model and no camera. It triggers the pipeline stages and views their results, and hosts the few steps that need a human in the loop. The main one is click-to-segment masking: a click in the UI calls a segmentation service (SAM~\citep{kirillov2023sam}) to produce the per-object masks that reconstruction and tracking consume.

\noindent\textbf{Pipeline.} The orchestration layer. Each step is a \emph{stage} (reconstruct scene, record demonstration, track objects, track hands, retarget) behind a uniform interface: read inputs from the database, call a service, write artifacts, record status. Because the stages are decoupled, a single stage can be run or re-run on its own, which is convenient for debugging or swapping one model.

\noindent\textbf{Services.} Each heavy model runs as a long-lived worker that loads its weights once and answers requests over a socket. A manager starts only the enabled workers, and a uniform client lets a stage call a model without knowing where it runs. Our motion planner (Sec.~\ref{sec:formulation}), the core of the retargeting solver, is one such worker: compiled once per scene and reused across its demonstrations, so the first solve has no cold-start cost.

\noindent\textbf{Data.} A database organizes the scenes, demonstrations, and their relationships, while bulk files stay on disk. This keeps the ``one scene, many demonstrations, one shared reconstruction'' structure explicit and easy to query.


\noindent\textbf{Sensing.} The raw appearance and geometry inputs, reused from the host stack and produced once upstream of all perception. A camera service provides the RGB/stereo stream, and a swappable stereo module turns it into metric depth, either \textsc{TRI-Stereo}~\citep{tristereo} or \textsc{FoundationStereo}~\citep{wen2025foundationstereo} as the backend.

\section{Design Choices}
\label{sec:design}

Each perception stage exposes a backend interface so its model can be swapped from config without modifying the pipeline. \textbf{Scene reconstruction} produces a watertight per-object mesh from a reference frame and a per-object mask~\citep{kirillov2023sam}, via \textsc{RecGen}~\citep{zadaianchuk2026recgen} or \textsc{SAM3D}~\citep{meta2025sam3d}. 
\textbf{Hand tracking} detects 2D hand landmarks with \textsc{MediaPipe Hands}~\citep{zhang2020mediapipe} or \textsc{HaMeR}~\citep{pavlakos2024hamer} and lifts them to a metric-scale 3D wrist/palm reference through the depth. \textbf{Object tracking} recovers per-frame 6D pose with \textsc{Any6D}~\citep{lee2025any6d} or \textsc{FoundationPose}~\citep{wen2024foundationpose}. 

\section{Retargeting Solver}
\label{sec:solver}

The solver works in three parts. \emph{Event extraction} recovers reliable contact events from the noisy tracks (Sec.~\ref{sec:events}). A \emph{coordinator} sequences the phases of a manipulation task and keeps the simulator in the loop to validate each phase and backtrack on failure (Sec.~\ref{sec:statemachine}). Every motion it plans goes through a collision-aware \emph{motion planner}, the algorithmic core, which we present on its own (Sec.~\ref{sec:formulation}).

\subsection{Event extraction}
\label{sec:events}

Event extraction recovers, from noisy single-view tracking, when the hand grasps and releases which object. We first clean the raw tracks with a noise filter: occlusion gaps are interpolated, spikes are removed with a median filter, and the hand keypoints and wrist pose are smoothed (Savitzky--Golay filter~\citep{savitzky1964smoothing}). Single-view keypoints are lifted to 3D through the depth map. Wherever the manipulated object foreground-occludes the hand, the depth is wrong, and so is the lifted joint. We denoise these by fitting a MANO hand model to the keypoints and re-reading clean, anatomically consistent joints (fingertips included) from the posed mesh. We then read contacts off \emph{kinematic co-motion}: an object is held over a window only if it moves together with the gripper while the gripper is moving. The output is a sequence of hold and idle states. The state transitions are defined as events, for a single arm, these states are rest, grasp, transport, and release. Each grasp event also records the human's hand pose at contact, a prior we use for grasp selection.

\subsection{Simulation-in-the-loop coordinator}
\label{sec:statemachine}

The coordinator is driven by a simple state machine with four phases: rest, grasp, transport, and release. It closes the simulation loop. At each phase it queries the planner with the current state, the phase's reference trajectories, the scene's collision, and which object is currently held (the
  attachment state). It runs the returned motion in the simulator and reads the outcome back. The simulator returns a settled state: the achieved configuration, the object poses, and what is currently held. This state is the coordinator's record of progress, and it seeds the next phase. A phase is committed only if the simulator confirms it: the gripper reaches and holds the object, the carried object follows its demonstrated motion, and neighbors stay put. Otherwise the coordinator backtracks: it restores the previous state and retries. The grasp phase, for example, retries with the next of a ranked set of candidate grasps (antipodal samples on the object, ordered by the demonstrated hand pose). Because planning and validation share one simulator, the committed run is at once a correctness check and a ready-to-use rollout.

\subsection{Motion planner}
\label{sec:formulation}

The motion planner is the algorithmic core of the solver. Retargeting a demonstration onto the arm poses two problems of different character. The first is global: the arm must route through clutter to a feasible grasp, choosing which way to pass each obstacle, a discrete decision a local search cannot undo. The second is local: along that route the motion must be smooth, dynamically feasible, and faithful to the demonstrated path while carrying any grasped object. We solve each with the tool suited to it. For the global route we reuse the demonstration when one is given, since a human demonstration is a good prior for a feasible trajectory, and fall back to a sampling-based planner otherwise; the local refinement is a trajectory optimizer, which doubles as the single collision-aware representation shared across all phases.

\textbf{Trajectory optimization.} Over a joint trajectory $\Q=(\q_1,\dots,\q_T)$ the optimizer minimizes a nonlinear least-squares cost on top of PyRoKi~\citep{kim2025pyroki,levenberg1944method}, with an augmented-Lagrangian wrapper promoting joint limits and a table stand-off to hard constraints,
\begin{equation}
\label{eq:obj}
J(\Q;\mathbf{m})\;=\;\underbrace{L_\mathrm{track}+L_\mathrm{attach}}_{\text{demonstration fidelity}}
  \;+\;\underbrace{L_\mathrm{smooth}+L_\mathrm{coll}+L_\mathrm{lim}}_{\text{feasibility}}.
\end{equation}
Here $\mathbf{m}$ is a vector of binary masks that specializes the objective to the current phase and attachment state. 
The terms in Eq.~(\ref{eq:obj}) have the following roles. 
$L_\mathrm{track}$ pulls the end-effector along the demonstrated reference path. 
$L_\mathrm{attach}$ keeps a held object rigidly attached to the gripper. 
$L_\mathrm{smooth}$ penalizes velocity and acceleration. 
$L_\mathrm{coll}$ penalizes self-collision and collision with active scene geometry. 
$L_\mathrm{lim}$ enforces joint limits.
The same masks also gate a fixed-size pool of collision primitives, so the sparsity pattern and compiled kernels remain unchanged across phases and objects. For an object that can be grasped, we include both world-frame and gripper-frame versions of its primitives; grasping the object masks out the world-frame version and masks in the gripper-frame version, rather than rebuilding the optimization problem.
The world-collision term keeps each sphere $s$ above an activation margin, penalizing $[\delta_s(t)-d_s(\q_t)]_+$, where $d_s(\q_t)$ is the signed distance from sphere $s$ to the active scene geometry at configuration $\q_t$. 
Here $\delta_s(t)$ is a clearance buffer that holds the arm off the scene; this is appropriate for obstacles but too conservative for the object being grasped, which the gripper must touch. 
We therefore taper it over the final $\tau$ knots of an approach for the grasp object alone,
\begin{equation}
\label{eq:taper}
\delta_s(t)=
\begin{cases}
\delta, & s\in\text{obstacles/floor},\\[2pt]
\delta\to\underline{\delta}_s, & s\in\text{grasp object},\ t>T-\tau,
\end{cases}
\end{equation}
so the gripper seats onto its target while avoidance around true obstacles is never relaxed.

\textbf{Global routing.} A descent of Eq.~(\ref{eq:obj}) is local: it cannot leave the homotopy class of its seed, because the buffered collision term gives no gradient toward the free gap once the seed already penetrates an obstacle. The global stage exists to place that seed in a collision-free homotopy class. It runs in three steps,
\begin{align}
\q^\star &= \arg\min_{\q}\ \big\|\,\mathrm{fk}(\q)\ominus x_T\big\|^2
   \quad\mathrm{s.t.}\ \ d_s(\q)\ge 0\ \ \forall s, \label{eq:seat}\\[2pt]
\Q^0 &= \begin{cases}
   \text{collision-aware sweep of the reference}, & \text{demo given},\\[2pt]
   \mathrm{interp}\big(\mathrm{RRT}(\q_0\!\rightarrow\!\q^\star)\big), & \text{otherwise},
   \end{cases} \label{eq:seed}\\[2pt]
\Q^\star &= \arg\min_{\Q}\ J(\Q;\mathbf{m})\ \ \text{seeded at}\ \Q^0, \label{eq:refine}
\end{align}
where $\mathrm{fk}(\q)$ is the gripper pose, $x_T$ the demonstrated terminal pose, $\ominus$ the $SE(3)$ error, and $\q_0$ the current configuration. The seat IK~(\ref{eq:seat}) drives the gripper to $x_T$ under hard no-penetration, so $\q^\star$ seats as deep onto the target as it can without colliding, identically for grasping or placing. 
%
The seed~(\ref{eq:seed}) is the global route: when a reference path is available, we use it as a homotopy prior and project it through the collision-aware model; otherwise RRT-Connect~\citep{kuffner2000rrtconnect} routes from $\q_0$ to $\q^\star$.
This route mainly lifts goal-reaching, where a single descent has no path to the seated goal and stalls in clutter (Sec.~\ref{sec:eval}). Either route is geometric, so the refinement~(\ref{eq:refine}) hands it to the optimizer, which smooths it, enforces the velocity and acceleration limits, and tracks the reference in one descent. The two are complementary: the route owns global feasibility, the optimizer owns smoothness, dynamics, and fidelity.

\textbf{Single-stage ablation.} To isolate the global stage, we report a single-stage variant that drops it and runs one descent of Eq.~(\ref{eq:obj}) from the current state. It is lighter, but being only local it degrades sharply in clutter, as Sec.~\ref{sec:eval} quantifies.

\section{Experiments}
\label{sec:eval}

DemoBridge builds a digital twin of a tabletop scene and retargets single-view human demonstrations onto a robot arm. Our evaluation has two parts. First, we isolate the motion planner and measure its tracking, endpoint reach, and collision behavior on a controlled synthetic benchmark. Second, we measure retargeting success on real demonstrations.

\subsection{Synthetic planner benchmark}

\begin{figure}[t]
\centering
\includegraphics[width=0.49\linewidth]{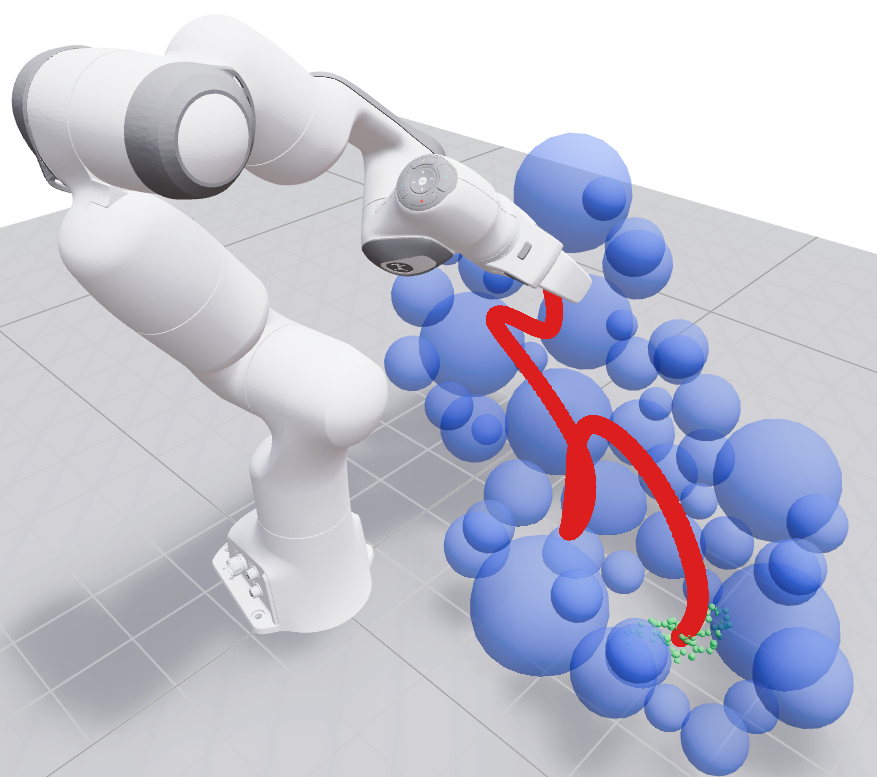}\hfill
\includegraphics[width=0.49\linewidth]{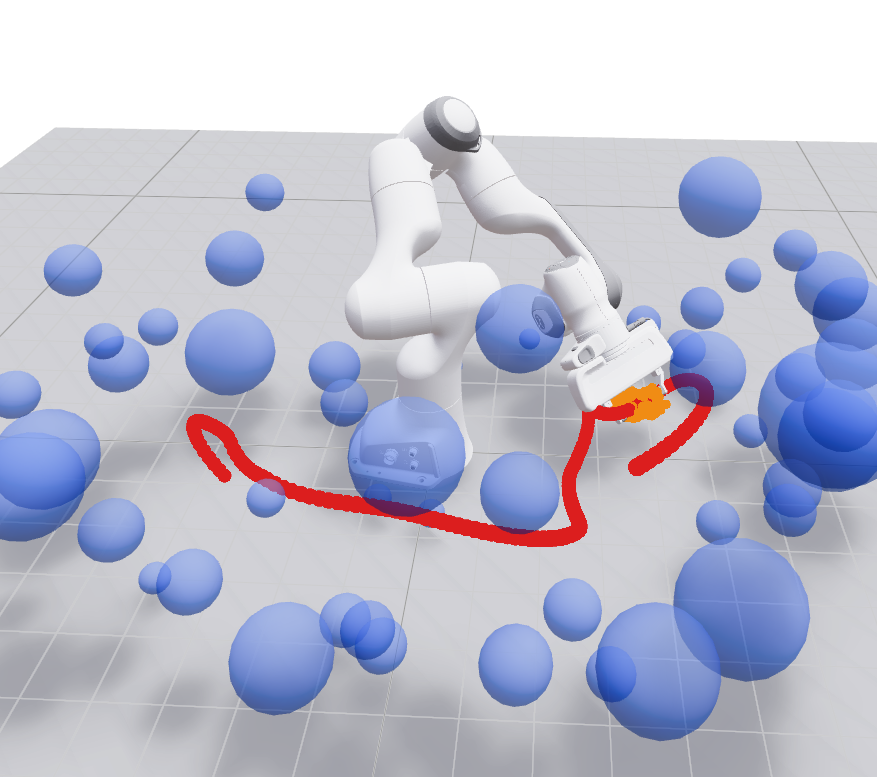}
\caption{Example generated synthetic scenes. Left: a grasp (base) scene, where the arm reaches the target object through surrounding clutter. Right: an attach scene, where the arm carries and places a grasped object along a round-trip path. The red curve is the reference end-effector path. The blue spheres denote sampled collision geometry used to approximate clutter obstacles in the planner.}
\label{fig:gen}
\end{figure}

\paragraph{Benchmark construction}
We construct a synthetic planner benchmark by procedurally generating cluttered tabletop manipulation scenes for a Franka arm, as shown in Fig.~\ref{fig:gen}. For each scene, we first specify a terminal end-effector pose corresponding to a valid grasp configuration. A target object is then instantiated between the gripper fingers at this terminal pose. The object is either a cuboid or a cylinder with a short-side width of 60mm and the long-side greater than gripper width, which is smaller than the gripper opening. To match the collision representation used by our real-world retargeting pipeline, we approximate the object surface with 100 spheres of radius 4mm. The benchmark has two versions. The base version evaluates reaching the target grasp. The attach Fig.~\ref{fig:gen} (right) version additionally attaches the grasped object to the gripper and requires the planner to carry it during the planning.

Given the terminal grasp pose, we generate a feasible robot trajectory in reverse. Starting from a region near the terminal pose, we repeatedly sample waypoints until their spatial distribution satisfies a prescribed coverage criterion. We then run motion planning between consecutive waypoints to obtain a continuous robot-arm trajectory. Reversing this trajectory, from the last sampled waypoint to the terminal grasp pose, gives a collision-free kinematic witness path for the robot. This reversal is used only as a benchmark construction device: it certifies geometric feasibility in a static scene, and does not assume that the underlying manipulation dynamics, such as gravity-driven object motion or failed contacts, are physically reversible.

After obtaining this witness trajectory, we record the swept volume of the robot arm. We then sample additional collision geometry only in regions at least 5cm away from this swept volume. 
This construction yields cluttered scenes that remain feasible by design: each scene is guaranteed to admit at least one collision-free robot trajectory, while the sampled obstacles still create challenging local constraints for planning. 
For the \emph{Attach} split in Table~\ref{tab:solver} we remove additional spheres that collide with the attached object. This results in a sparser collision environment. We report on a held-out evaluation set of 50 such scenes, generated with a fresh random seed.

\paragraph{Metrics}
We report three metrics. Endpoint error measures the Euclidean distance between the final gripper pose and the target grasp pose, averaged over the 50 scenes; a trial is considered successful if this error is below 5mm. Collision is evaluated by densely auditing the full trajectory with 32 interpolation substeps between trajectory knots and flagging any object penetration greater than 2mm. We report the number of scenes, out of 50, in which any collision occurs. Finally, when a reference path is provided, we report normalized dynamic-time-warping similarity, \emph{nDTW}, between the executed and reference end-effector paths, averaged over the scenes. The score lies in $(0,1]$, with 1 indicating an exact path match.

\begin{table*}[t]
\centering
\small
\setlength{\tabcolsep}{4.5pt}
\begin{tabular}{l cccc cccc}
\toprule
& \multicolumn{4}{c}{\emph{Base} (grasp object)} & \multicolumn{4}{c}{\emph{Attach} (carry and place)} \\
\cmidrule(lr){2-5}\cmidrule(lr){6-9}
Method & End.\ err.\ (mm) & Succ.\ (<5mm) & Coll. & nDTW & End.\ err.\ (mm) & Succ.\ (<5mm) & Coll. & nDTW \\
\midrule
\multicolumn{9}{l}{\emph{Goal only (no reference)}}\\
Trajopt                  & 9.1 & 26\% & 1/50 & -- & 10.1 & 12\% & 0/50 & -- \\
Single-stage             & 16.8 & 0\% & 30/50 & -- & 6.9 & 50\% & 2/50 & -- \\
\textbf{Multi-stage}     & \textbf{3.2} & \textbf{98\%} & 0/50 & -- & \textbf{3.3} & \textbf{98\%} & 0/50 & -- \\
\midrule
\multicolumn{9}{l}{\emph{With reference path}}\\
Trajopt                  & 11.8 & 12\% & 7/50 & 0.11 & 10.1 & 12\% & 0/50 & 0.08 \\
Trajopt (seg)            & 9.2 & 12\% & 7/50 & 0.56 & 8.7 & 18\% & 0/50 & 0.70 \\
Single-stage             & 18.9 & 0\% & 29/50 & 0.75 & 6.8 & 54\% & 1/50 & \textbf{0.93} \\
\textbf{Multi-stage}     & \textbf{4.6} & \textbf{68\%} & 1/50 & \textbf{0.78} & \textbf{3.0} & \textbf{100\%} & 0/50 & 0.86 \\
\bottomrule
\end{tabular}
\caption{Planner comparison on the held-out $50$-scene synthetic benchmark. Succ.\ is the fraction reaching within 5mm; Coll.\ counts scenes with any collision. Best per column in bold. End.\ err.\ reports the mean final gripper-position error over 50 scenes.}
\label{tab:solver}
\end{table*}

\paragraph{Evaluation settings}
These synthetic scenes isolate planning performance from perception noise. We evaluate our single-stage and multi-stage planners against PyRoKi trajectory optimization, denoted trajopt~\citep{kim2025pyroki}, under two settings, as shown in Table~\ref{tab:solver}. In the \emph{goal-only} setting, the planner is given only the terminal pose. This setting tests whether our method preserves standard goal-reaching performance when no reference path is available. In the \emph{with-reference} setting, the planner is additionally given the full demonstrated end-effector path. For this setting, we also evaluate a segmented trajopt baseline, denoted seg, which chains coarse reference waypoints using trajectory optimization. We run all settings on both the base and the attach version of the benchmark.

The two settings are different tasks and should not be compared against each other. Goal-only may reach the target by any collision-free route, whereas with-reference must largely follow the demonstrated path. Our scenes place obstacles off this path rather than across every approach, so easier routes to the goal remain open; following the reference is therefore the stricter task, and a lower score under with-reference reflects this added constraint rather than a regression. Each setting is meaningful only as a comparison across methods within its own column.

\paragraph{Results}
Table~\ref{tab:solver} reports the comparison, from which we read four findings.

\textbf{The multi-stage planner almost never collides, on either benchmark version and in either setting.} Across all four conditions it collides on at most one of the 50 scenes, including the attach version where the carried object must also stay clear. The single-stage ablation, sharing the same objective but no global plan, collides on 29 to 30 of 50 base scenes. Trajopt collides on up to 7 base scenes. Collision avoidance is therefore central to our pipeline, and the multi-stage planner is the only method that remains near-collision-free across all settings.

\textbf{Without a reference, our planner outperforms the initial Trajopt strategy.} In the goal-only setting, the multi-stage planner reaches an average endpoint error of about 3mm on both versions, well below trajopt at 9 to 10mm, while colliding on zero scenes. The extra machinery that retargeting needs costs nothing on standard goal-reaching.

\textbf{With a reference, only our planner follows it without colliding.} Trajopt reaches toward the goal but ignores the demonstrated path, at an nDTW of 0.11 on the base version, and the segmented variant raises this only to 0.56. The single-stage planner tracks the path more closely but breaks down in clutter, with 18.9mm endpoint error and 29 of 50 colliding scenes on the base version. The multi-stage planner holds endpoint error to 4.6mm on the base version and 3.0mm on the attach version with at most one collision, while keeping path fidelity high (nDTW 0.78 and 0.86). It is the only method that combines accurate reach, collision-freedom, and faithful path following.

\textbf{The global planning stage is what gives clutter robustness.} The single-stage ablation shares our objective and compiled kernels but runs a single local descent from one seed. The global planning stage that precedes the optimization finish is what removes the clutter collisions, by seeding the optimizer in a collision-free homotopy class it cannot reach on its own, and it does so without sacrificing endpoint accuracy or reference fidelity.

\begin{table}[t]
\centering
\small
\setlength{\tabcolsep}{4.5pt}
\begin{tabular}{llcccc}
\toprule
Task & Method & Reach & Grasp & Transp. & Place \\
\midrule
\multirow{3}{*}{Mug-on-plate} & YOTO$^\dagger$       & 6/10 & 0/10 & 0/10 & 0/10 \\
                              & RoboWheel$^\dagger$  & 7/10 & 3/10 & 3/10 & 2/10 \\
                              & Ours                 & 8/10 & 8/10 & 8/10 & 6/10 \\
\addlinespace
\multirow{3}{*}{Cup-stack}    & YOTO$^\dagger$       & 7/10 & 3/10 & 3/10 & 0/10 \\
                              & RoboWheel$^\dagger$  & 8/10 & 1/10 & 1/10 & 0/10 \\
                              & Ours                 & 9/10 & 9/10 & 9/10 & 7/10 \\
\addlinespace
\multirow{3}{*}{Pour-milk}    & YOTO$^\dagger$       & 8/10 & 1/10 & 1/10 & 1/10 \\
                              & RoboWheel$^\dagger$  & 10/10 & 1/10 & 1/10 & 1/10 \\
                              & Ours                 & 8/10 & 8/10 & 8/10 & 7/10 \\
\bottomrule
\end{tabular}
\caption{Real-demonstration success through four cumulative stages (of 10). YOTO$^\dagger$, RoboWheel$^\dagger$: our reimplementations of the closest methods.}
\label{tab:success}
\end{table}

\begin{figure*}[t]
\centering
\includegraphics[width=\textwidth]{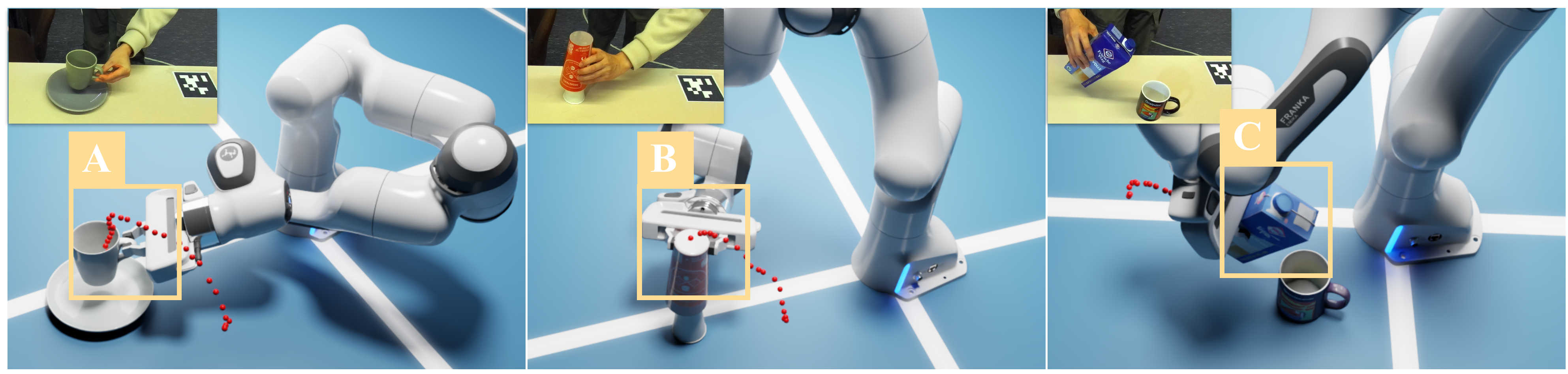}
\caption{Real-world retargeting results from single-view human demonstrations. \textbf{(A)}~mug-on-plate: the demonstrated hand trajectory suits the arm's reach, and the solver follows it directly. \textbf{(B)}~cup-stack: the demonstrated grasp pose is too close to the arm, so the solver seats the grasp from a different feasible approach direction. \textbf{(C)}~pour-milk: the retargeted arm trajectory is essentially correct, but an object-pose tracking error inverts the manipulated object's orientation in simulation, a perception limitation rather than a planner one.}
\label{fig:rollout}
\end{figure*}

\subsection{Real demonstrations}

\paragraph{Tasks}
We run the full pipeline on three tabletop scenes of increasing precision, with ten human demonstrations each. \textbf{Mug-on-plate} is a pick-and-place task with forgiving placement. \textbf{Cup-stack} places one cup onto a slightly smaller cup, so the final seating is precision-critical. \textbf{Pour-milk} is a trajectory-following task: the object must reproduce the demonstrated pouring orientation over the mug. Across all real-demonstration experiments, we use \textsc{RecGen}~\citep{zadaianchuk2026recgen} for scene reconstruction, \textsc{FoundationPose}~\citep{wen2024foundationpose} for object-pose tracking, \textsc{MediaPipe}~\cite{zhang2020mediapipe} for hand tracking, and the default multi-stage collision-aware planner for retargeting.

\paragraph{Metrics}
We evaluate each retargeted demonstration under full physics simulation and score its progress through four cumulative stages: \emph{reach}, \emph{grasp}, \emph{transport}, and \emph{place}. The reach stage is satisfied when the gripper approaches the object region. The grasp stage requires the gripper to pick up the object. The transport stage requires the carried object to follow the demonstrated motion while remaining clear of environment objects. The final place stage corresponds to task completion.

\paragraph{Baselines}
Neither of the closest methods ships with a retargeting pipeline we can run on our setup: RoboWheel's code is unreleased, and YOTO publishes only part of its extraction pipeline. We therefore reimplement the two recipes. YOTO$^\dagger$~\citep{zhou2025yoto} reduces a demonstration to sparse keyframes and reaches them by inverse-kinematics interpolation. RoboWheel$^\dagger$~\citep{zhang2025robowheel} tracks the dense end-effector path with per-frame inverse kinematics under self-collision only. Neither plans the whole arm against the scene, and neither has a fallback when a hand pose is infeasible for the arm. We cannot reproduce their grasp-timing extraction, so we manually annotate the grasp and release frames for both reimplementations. Our own solver is annotation-free: it is the default multi-stage planner of Sec.~\ref{sec:formulation} and infers timing from the hand--object relation (Sec.~\ref{sec:events}).

\paragraph{Results}
Table~\ref{tab:success} reports success through the four cumulative stages. The following findings summarize the takeaways.

\textbf{Grasp timing is necessary but far from sufficient.} We supply the baselines with the grasp timing they cannot reliably extract, yet both clear only the reach stage and rarely grasp. The reference pose is the deeper problem. From a single view it is accurate only to a few millimeters. When the object is nearly as wide as the gripper opening, that offset is enough to miss the grasp. The baselines track this reference directly, so they fail where re-seating the grasp would succeed.

\textbf{Pose-by-pose retargeting collides once the arm enters the scene.} YOTO$^\dagger$ interpolates between keyframes without checking the scene, so the arm or the carried object strikes a near-table obstacle or the support object before transport completes. RoboWheel$^\dagger$ tracks the dense path greedily, which drives the arm into configurations from which later poses are infeasible and triggers re-plans or IK failures. Both drop most rollouts before placement.

\textbf{Our planner carries most demonstrations to completion.} It grasps reliably and reaches the final place stage on a majority of demonstrations in every task, well above both baselines. The gain comes from optimizing against the whole demonstrated trajectory while staying collision-aware, rather than chasing individual poses.

\textbf{Failures arise where the feasible set is thin.} A demonstrated grasp fixes both a position and a gripper orientation. When that pose sits near the arm's reach limit, the feasible set of robot configurations is thin, and the non-convex optimizer can return a suboptimal solution. This is what separates the tasks. Mug-on-plate is the hardest: some demonstrated gripper orientations push the grasp into such a region, costing rollouts at the reach stage. Cup-stack, by contrast, keeps its grasps in well-conditioned configurations and stays reliable despite its precision-critical seating. Transport never fails once a rollout has grasped and begun to carry.

Fig.~\ref{fig:rollout} shows representative rollouts and failure modes. Fig.~\ref{fig:rollout}A shows a standard successful mug-on-plate execution. Fig.~\ref{fig:rollout}B illustrates a case where directly imitating the human reference pose would place the robot close to its base and induce self-collision. Instead, our retargeting solver finds an alternative feasible grasp configuration and still completes the task successfully. Fig.~\ref{fig:rollout}C highlights a limitation of the current pipeline: errors in object-pose tracking can propagate into retargeting. In this example, the estimated object pose is flipped, but the rollout still succeeds. In more challenging cases, limited image resolution, object occlusion, or motion blur can lead to more severe tracking failures. Across our demonstrations, tracking stays within the task tolerance. Thus, compared with perception, the success rates in Table~\ref{tab:success} primarily reflect retargeting quality.

\section{Limitations}
\label{sec:limits}

The main limitation is object tracking. In real use the hand partially occludes the object during manipulation, and the per-frame 6D pose tracker then often loses track of it. This corrupts the reference trajectory and the extracted events. A second limitation is bimanual support. The solver currently supports a single arm. The bimanual event extraction is already reliable with an additional handover event, but bimanual motion planning has not yet been included. Its modular design makes it easy to extend to new embodiments and to swap in stronger intermediate perception models as they become available.

\section*{Acknowledgment}
This work was funded by Toyota Motor Europe. We thank Andrii Zadaianchuk, Leonardo Barcellona, Christian Gumbsch, and Stratis Gavves for their valuable discussions.

\small
\bibliographystyle{plainnat}
\bibliography{references}

\end{document}